\pdfoutput=1

\documentclass[11pt]{article}

\usepackage[]{acl}

\usepackage{times}
\usepackage{latexsym}
\usepackage{amssymb}
\usepackage{bm}
\usepackage{hyperref}
\usepackage{color,colortbl}
\usepackage{stfloats}
\usepackage{enumitem}
\usepackage{booktabs}
\usepackage{nccmath}
\usepackage{multirow}
\usepackage{subfigure}
\usepackage{latexsym}
\usepackage{todonotes}
\usepackage{makecell}
\usepackage{dashrule}

\usepackage{graphicx}

\usepackage[T1]{fontenc}

\usepackage[utf8]{inputenc}

\usepackage{microtype}

\usepackage{soul}

\definecolor{LightGray}{gray}{0.9}
\usepackage{soul}

\title{Multilingual Pre-training with Language and Task Adaptation for Multilingual Text Style Transfer}

\author{
Huiyuan Lai, Antonio Toral, Malvina Nissim\\
CLCG, University of Groningen / The Netherlands\\
\texttt{\{h.lai, a.toral.ruiz, m.nissim\}@rug.nl}
}

\begin{document}
\maketitle
\begin{abstract}
We exploit the pre-trained seq2seq model mBART for multilingual text style transfer. Using machine translated data as well as gold aligned English sentences yields state-of-the-art results in the three target languages we consider.  Besides, in view of the general scarcity of parallel data, we propose a modular approach for multilingual formality transfer, which consists of two training strategies that target adaptation to both language and task. Our approach achieves competitive performance without monolingual task-specific parallel data and can be applied to other style transfer tasks as well as to other languages.
\end{abstract}

\section{Introduction}
\label{sec:intro}
Text style transfer (TST) is a text generation task where a given sentence must get rewritten changing its style  while preserving its meaning. 
Traditionally, tasks such as swapping the polarity of a sentence (e.g.\ ``This restaurant is getting worse and worse.''$\leftrightarrow$``This restaurant is getting better and better.'') as well as changing the formality of a text (e.g.\ ``it all depends on when ur ready.''$\leftrightarrow$``It all depends on when you are ready.'') are considered as instances of TST. We focus here on the latter case only, i.e.\ \textit{formality transfer}, because (i) recent work has shown that polarity swap is less of a style transfer task, since meaning is altered in the transformation~\citep{lai2021generic}, and (ii) data in multiple languages has recently become available for formality transfer~\citep{briakou-etal-2021-ola}.

Indeed, mostly due to the availability of parallel training and evaluation data, almost all prior TST work focuses on monolingual (English) text~\citep{rao-tetreault-2018, li-etal-2018, prabhumoye-etal-2018-style, cao-etal-2020-expertise}.\footnote{``Parallel data'' in this paper refers to sentence pairs in the same language, with the same content but different formality.} As a first step towards multilingual style transfer,~\citet{briakou-etal-2021-ola} have  released XFORMAL, a benchmark of multiple formal reformulations of informal text in Brazilian Portuguese (BR-PT), French (FR), and Italian (IT).
For these languages the authors have manually created evaluation datasets. On these, they test several monolingual TST baseline models developed 
using language-specific pairs obtained by machine translating GYAFC, 
a English corpus for formality transfer~\cite{rao-tetreault-2018}. \citet{briakou-etal-2021-ola} find that the models trained on translated parallel data do not outperform a simple rule-based system based on handcrafted transformations, especially on content preservation, and conclude that formality transfer on languages other than English is particularly challenging.

One reason for the poor performance could be the low quality (observed upon our own manual inspection) of the pseudo-parallel data, especially the informal side. Since machine translation systems are usually trained with formal texts like news~\citep{zhang-etal-2020-parallel}, informal texts are harder to translate, or might end up  more formal when translated.  
But most importantly, the neural models developed by \citet{briakou-etal-2021-ola} do not take advantage of two recent findings: (i) pre-trained models, especially the sequence-to-sequence model BART~\citep{lewis-etal-2020-bart}, have proved to help substantially with content preservation in style transfer~\citep{lai2021formality}; (ii) Multilingual Neural Machine Translation~\citep{johnson-etal-2017-googles, aharoni-etal-2019-massively, liu-etal-2020-multilingual-denoising} and Multilingual Text Summarization~\citep{hasan-etal-2021-xl} have achieved impressive results leveraging multilingual models which allow for cross-lingual knowledge transfer.

In this work we use the multilingual large model mBART~\citep{liu-etal-2020-multilingual-denoising} to model style transfer in a multilingual fashion exploiting available parallel data of one language (English) to transfer the task and domain knowledge to other target languages. 
To address real-occurring situations, in our experiments we also simulate complete lack of parallel data for a target language (even machine translated), and lack of style-related data at all (though availability of out-of-domain data). Language specificities are addressed through adapter-based strategies~\citep{pfeiffer-etal-2020-mad, ustun-etal-2020-udapter, ahmet2021multilingual}.
We obtain state-of-the-art results in all three target languages
, and propose a modular methodology that can be applied to other style transfer tasks as well as to other languages. We release our code and hopefully foster the research progress.\footnote{All code at \url{https://github.com/laihuiyuan/multilingual-tst}.}

\section{Approach and Data}
\label{sec:data}

As a base experiment aimed at exploring the contribution of mBART~\citep{liu-etal-2020-multilingual-denoising, tang2020multilingual} for multilingual style transfer, we fine-tune this model with parallel data specifically developed for style transfer in English (original) and three other languages (machine translated).

Next, in view of the common situation where parallel data for a target language is not available, we propose a two-step adaptation training approach on mBART that enables modular multilingual TST. We avoid iterative back-translation (IBT)~\citep{hoang-etal-2018-iterative}, often used in previous TST work~\citep{prabhumoye-etal-2018-style, lample2019multipleattribute, xiaoyuan-ijcai, lai2021generic}, since it has been shown to be computationally costly \citep{ahmet2021multilingual, stickland2021multilingual}. We still run comparison models that use it.

In the first adaptation step, we address the problem of some languages being not well represented in mBART, which preliminary experiments have shown to hurt our downstream task.\footnote{The number of monolingual sentences used in mBART-50's pre-training is only 49,446 for Portuguese, for example, versus 36,797,950 for French and 226,457 for Italian.} We conduct a language adaptation denoising training using unlabelled data for the target language. 
In the second step, we address the task at hand through fine-tuning cross-attention with auxiliary gold parallel English data adapting the model to the TST task.

For TST fine-tuning, we use parallel training data, namely formal/informal aligned sentences (both manually produced for English and machine translated for three other languages). For the adaptation strategies, we also collect formality and generic non-parallel data. Details follow.  

\paragraph{English formality data} GYAFC~\citep{rao-tetreault-2018} is an English dataset of aligned formal and informal sentences. Gold parallel pairs are provided for training, validation, and test.

\paragraph{Multilingual formality data} XFORMAL~\citep{briakou-etal-2021-ola} is a benchmark for multilingual formality transfer, which provides an evaluation set that consists of four formal rewrites of informal sentences in BR-PT, FR, and IT. This dataset contains pseudo-parallel corpora in each language, obtained via machine translating the English GYAFC pairs. 

\paragraph{Language-specific formality non-parallel data} Following~\citet{rao-tetreault-2018} and~\citet{briakou-etal-2021-ola}, we crawl the domain data in target language from Yahoo Answers.\footnote{\url{https://webscope.sandbox.yahoo.com/catalog.php?datatype=l&did=11}}

We then use the style regressor from~\citet{briakou-etal-2021-evaluating} to predict formality score $\sigma$ of the sentence to automatically select sentences in each style direction.\footnote{Sentences with $\sigma < -0.5$ are considered informal while $> 1.0$ are formal in our experiments.} 

\paragraph{Language-specific generic non-parallel data} 5~M sentences containing 5 to 30 words for each language randomly selected from News Crawl.\footnote{\url{ http://data.statmt.org/news-crawl/}}

\begin{figure*}[!t]
    \centering
    \subfigure[Language adaptation training with monolingual data]{
       \includegraphics[scale=0.47]{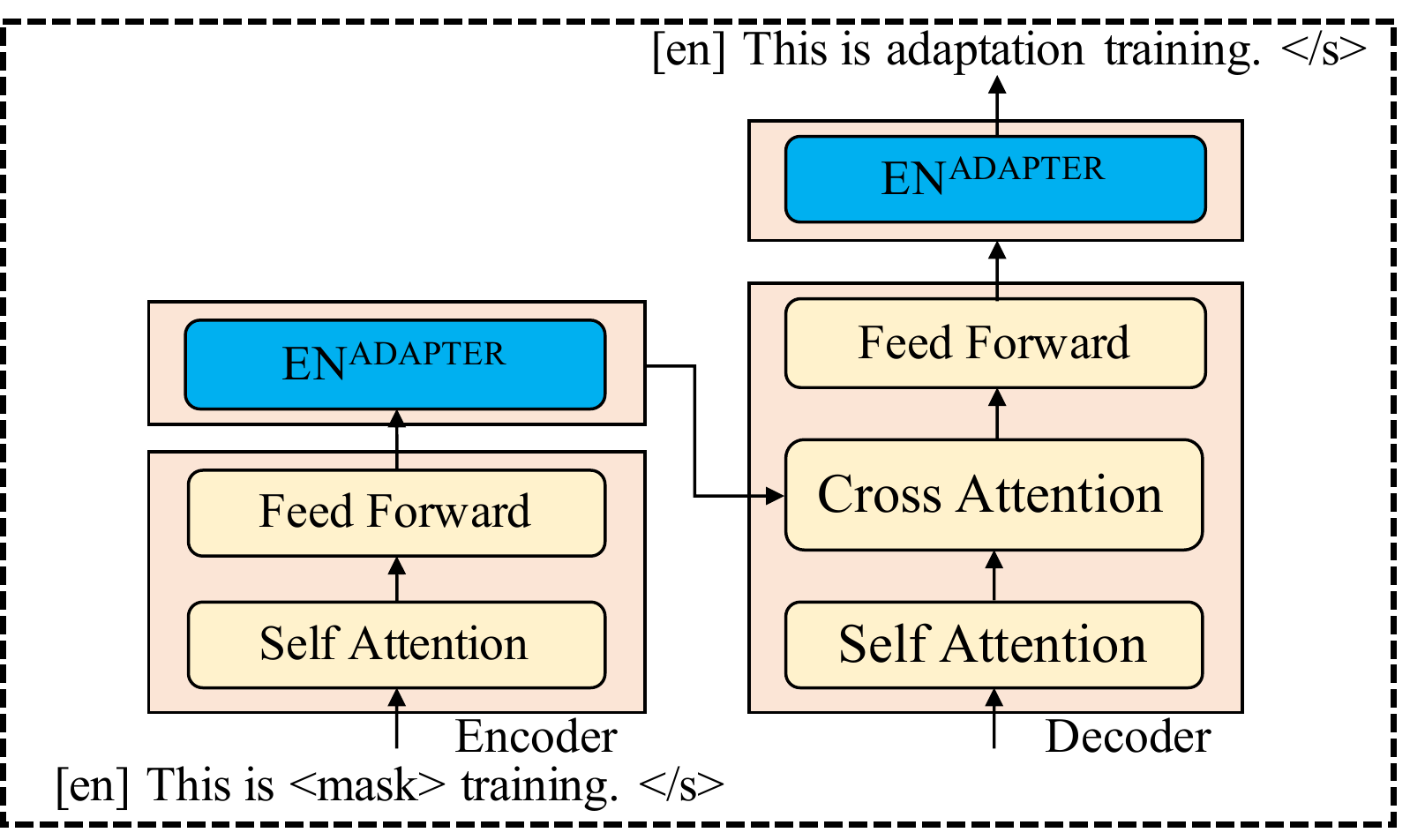}
       \label{fig:lang-adap}
    }
    \subfigure[Task adaptation training with English parallel data]{
        \includegraphics[scale=0.47]{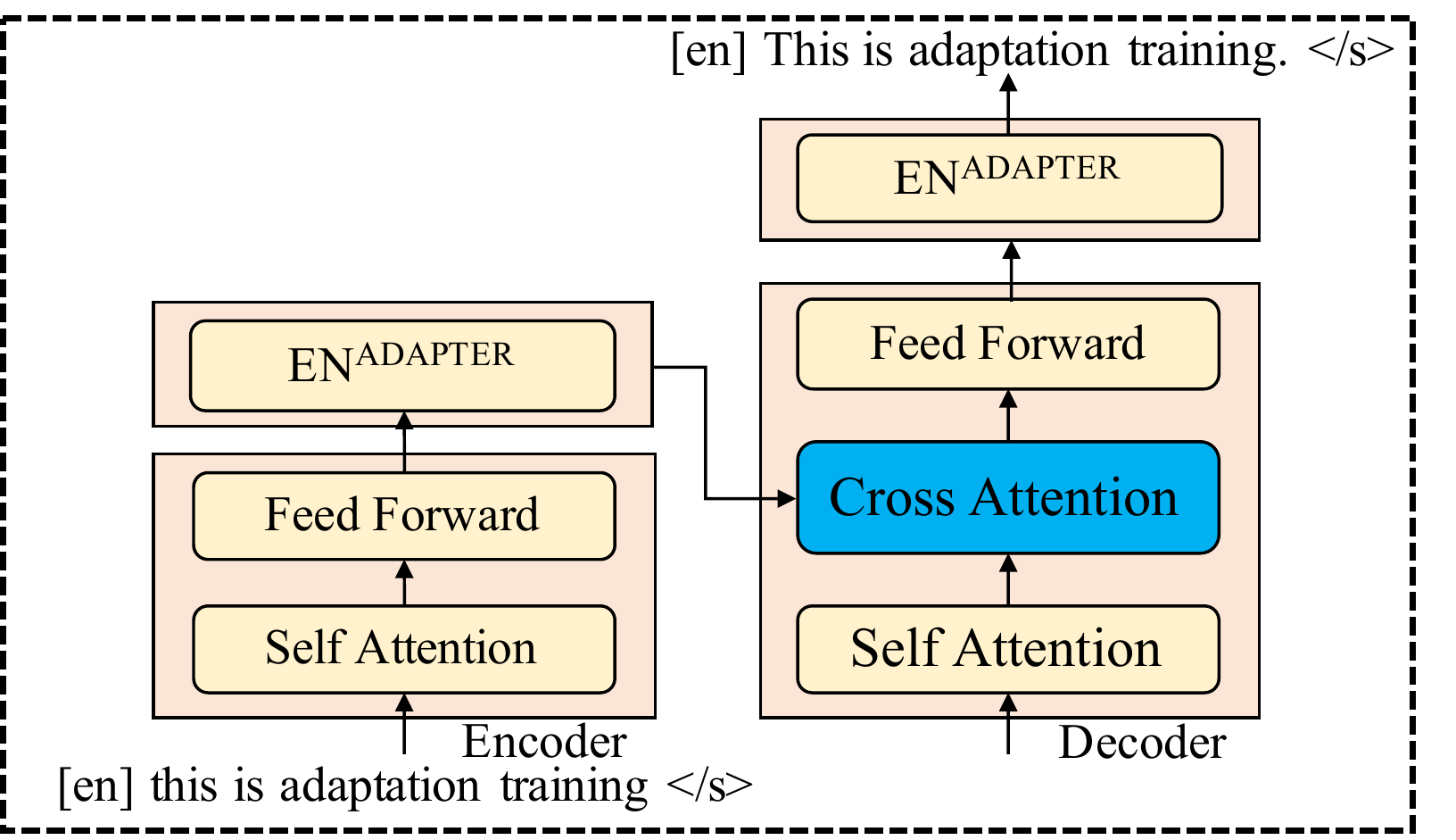} 
        \label{fig:task-adap}
    }
    \caption{Overview of adaptation training. In~\ref{fig:lang-adap}, the feed-forward network of each transformer layer or the inserted adapter layer is trained with monolingual data to adapt to the target language. In~\ref{fig:task-adap}, the cross-attention of mBART is trained with auxiliary English parallel data to adapt to the TST task.}
    \label{fig:evaluations}
\end{figure*}

\section{Adaptation Training}
\label{sec:adaptation}

To adapt mBART to multilingual TST, we employ two adaptation training strategies that target language and task respectively.

\subsection{Language Adaptation}
As shown in Figure~\ref{fig:lang-adap}, we introduce a module for language adaptation.
Inspired by previous work~\citep{houlsby2019parameter, bapna-firat-2019-simple}, we use an adapter 
(ADAPT; \textasciitilde50M parameters), which is inserted into each layer of the Transformer encoder and decoder, after the feed-forward block.

Following~\citet{bapna-firat-2019-simple}, the ADAPT module $A_{i}$ at layer $i$ consists of a layer-normalization LN of the input $x_{i} \in \mathbb R^{h}$ followed by a down-projection $W_{down} \in \mathbb R^{h \times h}$, a non-linearity and an up-projection $W_{up} \in \mathbb R^{h \times h}$ combined with a residual connection with the input $x_{i}$:

\begin{equation}
\label{eq:adapter}
    A(x_{i}) = W_{up}\textrm{RELU}(W_{down}\textrm{LN}(x_{i})) + x_{i}
\end{equation}

\paragraph{Language adaptation training}
Following mBART's pretraining, we conduct the language adaptation training on a denoising task, which aims to reconstruct text from a corrupted version:
\begin{equation}
\label{eq:denoising}
    L_{\phi_{\boldsymbol{A}}} = -\sum \log (T \mid g(T); \phi_{\boldsymbol{A}})
\end{equation}

\noindent where $\phi_{\bm{A}}$ are the parameters of adaptation module $A$, $T$ is a sentence in target language and $g$ is the noise function that masks 30\% of the words in the sentence. 
Each language has its own separate adaptation module.
During language adaptation training, the parameters of the adaptation module are updated while the other parameters 
stay frozen.

\subsection{Task Adaptation}
As shown in Figure~\ref{fig:task-adap}, after training the language adaptation module we fine-tune the model on the auxiliary English parallel data with the aim of making the model adapt to the specific task of formality transfer. Following~\citet{cooper-stickland-etal-2021-recipes}, we only update the parameters of the decoder’s cross-attention (i.e.\ task adaptation module) while the other parameters are fixed, 
thus limiting computational cost and catastrophic forgetting.

\paragraph{Multilingual TST process}
For the language adaptation modules we have two settings: (i) adaptation modules \textbf{A}$^{E}_{s}$ on the encoder come from the model trained with source style texts, and modules \textbf{A}$^{D}_{t}$ on the decoder come from the model trained with target style texts (M2.X, Table~\ref{tab:auto-results}); (ii) both \textbf{A}$^{E}$ and \textbf{A}$^{D}$ are from a model trained with generic texts (M3.X), so there are no source and target styles for the adaptation modules.
For the task adaptation modules, we also have two settings: (i) the module is from the English model (X + EN cross-attn); (ii) fine-tuning the model of the target language with English parallel data (X + EN data).

\section{Experiments}

All experiments are implemented atop Transformers~\citep{wolf-etal-2020-transformers} using mBART-large-50~\citep{tang2020multilingual}.  
We train the model using the Adam optimiser~\citep{kingma2017adam} with learning rate 1e-5 for all experiments. We train the language adaptation modules with generic texts separately for each language for 200k training steps 
with batch size 
32, accumulating gradients over 8 update steps, and set it to 1 for other training.

\paragraph{Evaluation}
Following previous work \cite{fuli-2019, Abhilasha-2020}, we assess style strength and content preservation. We fine-tune mBERT~\citep{devlin-etal-2019-bert} with \citet{briakou-etal-2021-ola}'s pseudo-parallel corpora to evaluate the style accuracy of the outputs. We also use a style regressor from~\citet{briakou-etal-2021-evaluating}, which is based on XLM-R~\citep{conneau-etal-2020-unsupervised} and is shown to correlate well with human judgments.\footnote{Results of classifiers/regressor are in Appendix~\ref{app:class-regr}.} We calculate BLEU and COMET~\citep{rei-etal-2020-comet} to assess content preservation. 
As overall score, following previous work, we compute the harmonic mean (HM) of style accuracy and BLEU.

\paragraph{Systems}
\label{sec:systems}

Based on our data (Section~\ref{sec:data}), we have four settings for our systems. \textbf{D1}: pseudo-parallel data in the target language via machine translating the English resource; \textbf{D2}: non-parallel style data in the target language; \textbf{D3}: no style data in the target language; \textbf{D4}: no parallel data at all. The first three settings all contain gold English parallel data.

\begin{table*}[ht]
\centering
\resizebox{\linewidth}{!}{%
\begin{tabular}{c|l|ccc|ccc|ccc||ccc|ccc|ccc}
\toprule[1.5pt]
  & & \multicolumn{9}{c||}{\sc Informal$\rightarrow$Formal} & \multicolumn{9}{c}{\sc Formal$\rightarrow$Informal}\\
  \cline{3-20}
  \makecell[c]{\multirow{1}{*}{\sc Data}} & \makecell[c]{\multirow{1}{*}{\sc Model}} & \multicolumn{3}{c|}{\textsc{Italian}} & \multicolumn{3}{c|}{\textsc{French}} & \multicolumn{3}{c||}{\textsc{Portuguese}} & \multicolumn{3}{c|}{\textsc{Italian}} & \multicolumn{3}{c|}{\textsc{French}} & \multicolumn{3}{c}{\textsc{Portuguese}}\\
  \cline{3-20}
  & & BLEU & ACC & HM & BLEU & ACC & HM & BLEU & ACC & HM & BLEU & ACC & HM & BLEU & ACC & HM & BLEU & ACC & HM \\

 \midrule
  \multirow{3}{*}{D1} 
  & Multi-Task~\citep{briakou-etal-2021-ola}  & 0.426 & 0.727 & 0.537 & 0.480 & 0.742 & 0.583 & 0.550 & 0.782 & 0.645 & - & - & - & - & - & - & - & - & -\\
  & M1.1: pseudo-parallel data                        & 0.459 & \underline{\textbf{0.856}} & \underline{\textbf{0.598}} & \textbf{0.530} & 0.829 & 0.647 & 0.524 & \textbf{0.852} & 0.649 & 0.177 & 0.311 & 0.226 & \textbf{0.195} & 0.377 & 0.257 & \textbf{0.225} & 0.306 & \textbf{0.259}\\
  & M1.2: M1.1 + EN data                     & \textbf{0.461} & 0.841 & 0.596 & 0.525 & \underline{\textbf{0.863}} & \underline{\textbf{0.653}} & \underline{\textbf{0.553}} & 0.809 & \underline{\textbf{0.657}} & \textbf{0.178} & \textbf{0.315} & \textbf{0.227} & 0.194 & \textbf{0.458} & \textbf{0.273} & 0.219 & \textbf{0.313} & 0.258\\

  \midrule
  \multirow{4}{*}{D2} & DLSM~\citep{briakou-etal-2021-ola}                  & 0.124 & 0.223 & 0.159 & 0.180 & 0.152 & 0.165 & 0.185 & 0.191 & 0.188 & - & - & - & - & - & - & - & - & -\\
  & M2.1: IBT training + EN data                  & 0.460 & 0.510 & 0.484 & 0.500 & 0.487 & 0.492 & 0.491 & 0.428 & 0.457 & 0.168 & 0.420 & 0.240 & 0.196 & 0.235 & 0.214 & \underline{\textbf{0.237}} & 0.083 & 0.123 \\
  & M2.2: ADAPT + EN cross-attn  & 0.467 & 0.637 & 0.539 & 0.516 & 0.627 & 0.566 & 0.499 & 0.365 & 0.422 & 0.175 & 0.672 & 0.278 & \textbf{0.212} & \textbf{0.627} & \textbf{0.317} & \underline{\textbf{0.237}} & 0.471 & \underline{\textbf{0.315}}\\
  & M2.3: ADAPT + EN data                & \textbf{0.476} & \textbf{0.731} & \textbf{0.577} & \textbf{0.519} & \textbf{0.702} & \textbf{0.597} & \textbf{0.526} & \textbf{0.509} & \textbf{0.517} & \textbf{0.180} & \textbf{0.719} & \textbf{0.288} & 0.209 & 0.567 & 0.305 & 0.169 & \textbf{0.534} & 0.257\\

  \midrule
  \multirow{3}{*}{D3} & M3.1: EN data     & \underline{\textbf{0.485}} & 0.670 & \textbf{0.563} & \underline{\textbf{0.553}} & 0.727 & 0.628 & 0.039 & \underline{\textbf{0.890}} & 0.074 & \underline{\textbf{0.186}} & \underline{\textbf{0.767}} & \underline{\textbf{0.299}} & \underline{\textbf{0.216}} & \underline{\textbf{0.692}} & \underline{\textbf{0.329}} & 0.020 & 0.403 & 0.038\\
  & M3.2: ADAPT + EN cross-attn  & 0.480 & 0.672 & 0.560 & 0.545 & \textbf{0.749} & \textbf{0.631} & \textbf{0.547} & 0.559 & \textbf{0.553} & 0.179 & 0.421 & 0.251 & 0.209 & 0.685 & 0.320 & 0.175 & \textbf{0.560} & 0.267\\
  & M3.3: ADAPT + EN data                & 0.423 & \textbf{0.735} & 0.537 & 0.547 & 0.722 & 0.622 & 0.423 & 0.508 & 0.462 & 0.169 & 0.733 & 0.275 & 0.205 & 0.584 & 0.303 & \textbf{0.189} & 0.505 & \textbf{0.275}\\

  \midrule
  \multirow{3}{*}{D4} 
  & Rule-based~\citep{briakou-etal-2021-ola} & \textbf{0.438} & \textbf{0.268} & \textbf{0.333} & \textbf{0.472} & \textbf{0.208} & \textbf{0.289} & \textbf{0.535} & \textbf{0.448} & \textbf{0.488} & - & - & - & - & - & - & - & - & -\\
  & M4.1: original mBART         & 0.380 & 0.103 & 0.162 & 0.425 & 0.080 & 0.135 & 0.128 & 0.200 & 0.156 & 0.160 & \textbf{0.146} & \textbf{0.153} & 0.189 & \textbf{0.189} & \textbf{0.189} & 0.080 & \underline{\textbf{0.657}} & \textbf{0.143}\\
  & M4.2: ADAPT (generic data)  & 0.401 & 0.092 & 0.150 & 0.444 & 0.075 & 0.128 & 0.463 & 0.223 & 0.301 & \textbf{0.164} & 0.130 & 0.145 & \textbf{0.194} & 0.170 & 0.181 & \underline{\textbf{0.237}} & 0.082 & 0.122\\
 \bottomrule[1.5pt]
\end{tabular}}
\caption{\label{tab:auto-results}
Results for multilingual formality transfer. Notes: (i) for F$\rightarrow$I there are four different source sentences and a human reference only, so for each instance scores are averaged; (ii) bold numbers denote best systems for each block, and underlined denote the best score for each transfer direction for each language.
}
\end{table*}

\paragraph{Results}
Table~\ref{tab:auto-results} shows the results for both I$\rightarrow$F (informal-to-formal) and F$\rightarrow$I (formal-to-informal) transformations.\footnote{Complete results are in Appendix~\ref{app:all-results}.} 
We include the models from~\citet{briakou-etal-2021-ola} for comparison (they only model the I$\rightarrow$F direction).

Results in \textbf{D1} show that fine-tuning mBART with pseudo-parallel data yields the best overall performance in the I$\rightarrow$F direction. The F$\rightarrow$I results, instead, are rather poor and on Italian even worse than IBT-based models (M2.1). This could be due to this direction being harder in general, since there is more variation in informal texts, but it could also be made worse by the bad quality of the informal counterpart in the translated pairs. 
Indeed, work in machine translation has shown that  low-quality data is more problematic in the target side 
than in the source side
~\citep{bogoychev2019domain}.

In \textbf{D2}, we see that our proposed adaptation approaches outperform IBT-based models on both transfer directions. The results of fine-tuning the target language's model with English parallel data are generally better than inserting the EN model's cross-attention module into the target language's model. This suggests that the former can better transfer task and domain knowledge.

In \textbf{D3}, the large amounts of generic texts yield more improvement in I$\rightarrow$F direction rather than F$\rightarrow$I. This could be due to generic texts being more formal than informal. The performance improvement on Portuguese is particularly noticeable (compare M3.1 trained with EN data only with other M3.X models), and mostly due to this language being less represented than the others in mBART. Interestingly, the performance of task adaptation strategies is reversed compared to D2: it is here better to adapt cross attention in the English model rather than fine-tune the target language model directly. Future work will need to investigate how using different data sources for language adaptation (D2, style-specific vs D3, generic) interacts with task adaptation strategies.

Results for \textbf{D4} show that language adaptation training helps with content preservation, especially for Portuguese, confirming this curbs the problem of language underrepresentation in pre-training. However, low performance on style accuracy shows that task-specific data is necessary, even if it comes from a different language.

\section{Analysis and Discussion}

\paragraph{Case Study} Table~\ref{tab:example-ouptus} shows a group of example outputs in Italian. In the I$\rightarrow$F direction, most systems tend to copy a lot from the source and change formality words slightly. DLSM and Rule-based systems fail to transfer the formality style while others are successful to some extent: our M1.1 yields the best performance on the style strength. When looking at content, most outputs contain more or less part of the source sentence; Multi-Task system achieves the highest BLEU score but our systems (except for M3.3) have higher COMET scores, with M3.1 achieving the highest score. For the F$\rightarrow$I direction, we can see that M1.1 has the worst performance on style strength (its output is almost identical to the source), while M2.1, M3.1 and M3.2 generate the same output with the lowest regression score. Overall, M3.3 achieves the best performance on style and content.

\begin{table*}[!ht]
\centering
\resizebox{\linewidth}{!}{%
\begin{tabular}{lp{18cm}cccc}
\toprule
 \sc Model & \sc Sentence & REG. & ACC & BLEU & COMET\\
 \hline
 \rowcolor{LightGray}
 \multicolumn{6}{c}{\sc Informal$\rightarrow$Formal}\\ 
 \hline
 Source & se te ne vai secondo me e segno di debolezza e di paura se hai tanti problemi qui cerca di risolverli & - & - & - & - \\ & \textit{if you go away I think it's a sign of weakness and fear if you have many problems here try to solve them}\\
 Reference & Secondo il mio parere, il tuo andartene denota debolezza e paura, poiché se hai molti problemi, è necessario risolverli. & - & - & - & - \\
 & \textit{In my opinion, your going away denotes weakness and fear, since if you have many problems it is crucial to solve them.}\\
 \hline
 Multi-Task  & Se te ne vai secondo me e segno di debolezza e di paura, se hai molti problemi qui, cerca di risolverli. & 0.120 & 0.959 & \textbf{0.151} & 0.175\\
 DLSM        & Se te ne vai qualcosa e stesso di cui e di peggio se hai messo due soldi <unk> tutti i <unk> di <unk> & -2.666 & 0.014 & 0.015 & -1.563\\
 Rule-based  & Se te ne vai secondo me e segno di debolezza e di paura se hai tanti problemi qui cerca di risolverli & -1.340 & 0.430 & 0.029 & 0.423\\
 M1.1 & Secondo me, è segno di debolezza e di paura. Se hai tanti problemi qui, cerca di risolverli. & \textbf{0.742} & \textbf{0.995} & 0.035 & 0.658\\
 M2.1 & Se te ne vai secondo me e segno di debolezza e di paura. Se hai tanti problemi qui cerca di risolverli. & -0.243 & 0.978 & 0.028 & 0.634\\
 M3.1 & Se te ne vai, secondo me è segno di debolezza e di paura. Se hai tanti problemi, cerca di risolvere i problemi. & 0.310 & 0.992 & 0.026 & \textbf{0.728}\\
 M3.2 & Se te ne vai è segno di debolezza e di paura, se hai tanti problemi qui cerca di risolverli. & -0.225 & 0.971 & 0.037 & 0.639\\
 M3.3 & Its segno di debolezza e paura, se hai tanti problemi qui cerca di risolvere. & -0.092 & 0.692 & 0.126 & -0.968\\
 \hline
 \rowcolor{LightGray}
 \multicolumn{6}{c}{\sc Formal$\rightarrow$Informal}\\
 \hline
 Source & Se scrivi in italiano corretto avrai più possibilità di ricevere una risposta.& - & - & - & -\\
 & \textit{If you write in correct Italian you will have a better chance of receiving an answer.}\\
 Reference & se magari scrivi in italiano riusciamo a risponderti!!!& - & - & - & -\\
 & \textit{maybe if you write in Italian we can answer you !!!}\\
 \hline
 M1.1 & Se scrivi in italiano correttamente, avrai più possibilità di ottenere una risposta. & 1.580 & 0.001 & 0.071 & \textbf{0.566}\\
 M2.1 & se scrivi in italiano corretto avrai più possibilità di ricevere una risposta. & \textbf{0.221} & 0.896 & 0.083 & 0.557\\
 M3.1 & se scrivi in italiano corretto avrai più possibilità di ricevere una risposta. & \textbf{0.221} & 0.796 & 0.083 & 0.557\\
 M3.2 & se scrivi in italiano corretto avrai più possibilità di ricevere una risposta. & \textbf{0.221} & 0.796 & 0.083 & 0.557\\
 M3.3 & scrivi in italiano e avrai più possibilità di ricevere una risposta. & 0.891 & \textbf{0.878} & \textbf{0.084} & \textbf{0.566}\\
 
\bottomrule
\end{tabular}}
\caption{\label{tab:example-ouptus}
Example outputs in Italian and their sentence-level evaluation scores. Notes: (i) REG. indicates the score of the style regressor; (ii) ACC is the style confidence from the style classifier. 
}
\end{table*}

\begin{figure}[t]
    \centering
    \includegraphics[scale=0.4]{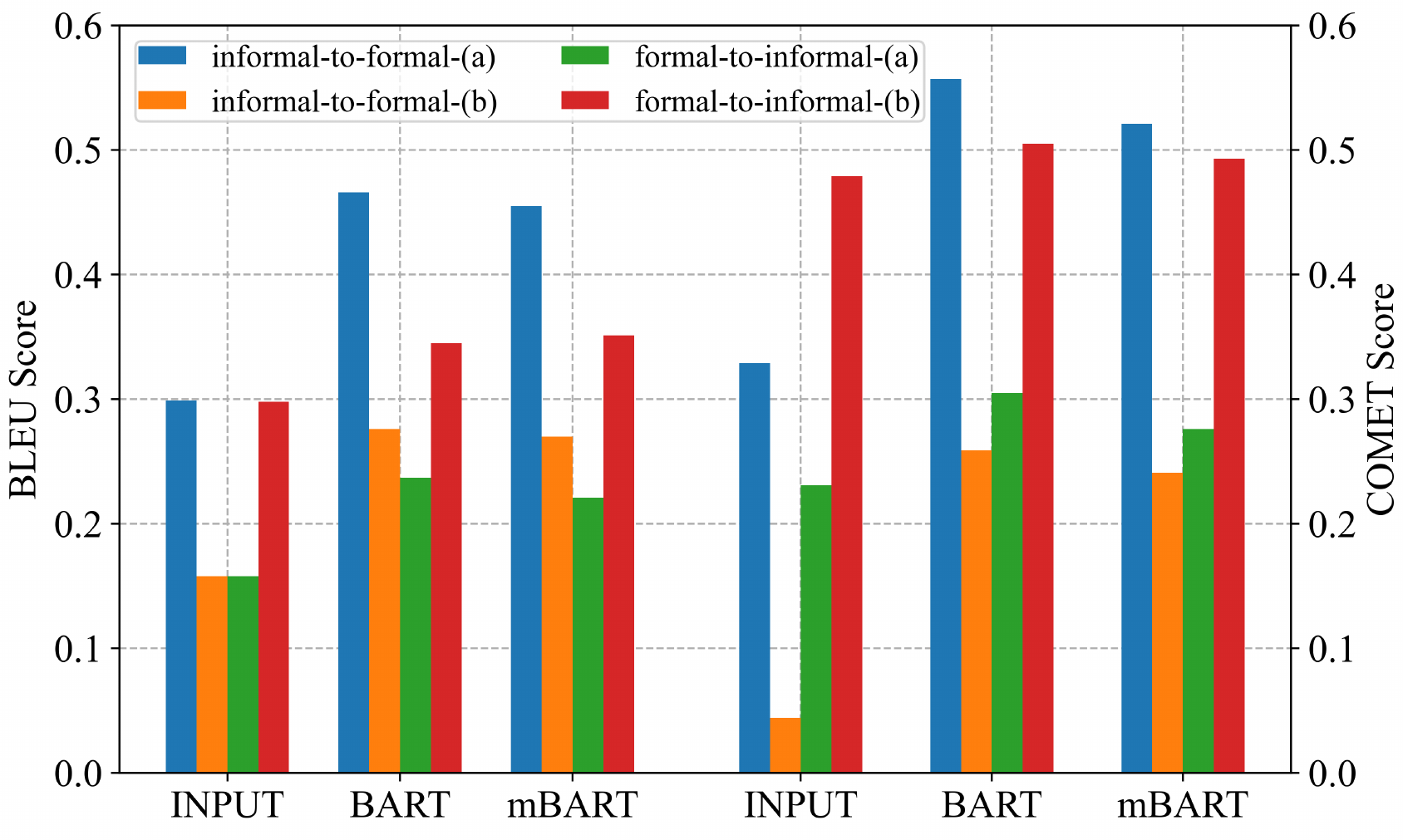}
    \caption{English formality transfer on content preservation using one reference. Setting (a) uses the original test set for each direction; (b) uses the test set of the opposite direction, swapping sources and references.
}
    \label{fig:en-content}
\end{figure}

\begin{table}[t]
\centering
\resizebox{\linewidth}{!}{%
\begin{tabular}{l|cc|cc|cc}
\toprule
 \makecell[c]{\multirow{3}{*}{\sc Model}} &  \multicolumn{2}{c|}{\sc Italian} &  \multicolumn{2}{c|}{\sc French} &  \multicolumn{2}{c}{\sc Portuguese}\\
 \cline{2-7}
 & BLEU & COMET & BLEU & COMET & BLEU & COMET\\
 \cline{2-7}
 & \multicolumn{6}{c}{\textsc{Informal$\rightarrow$Formal} (setting (a))}\\
 \midrule
 INPUT     & 0.176 & 0.078 & 0.198 & -0.019 & 0.244 & 0.217\\
 M1.1      & 0.196 & 0.170 & 0.234 & 0.133 & 0.269 & 0.282\\
 M1.2      & 0.194 & 0.181 & 0.231 & 0.138 & 0.283 & 0.319\\
 \toprule[0.8pt]
 & \multicolumn{6}{c}{\textsc{Formal$\rightarrow$Informal} (setting (b))}\\
 \midrule
 INPUT     & 0.174 & 0.364 & 0.196 & 0.277 & 0.243 & 0.463\\
 M1.1      & 0.194 & 0.326 & 0.201 & 0.239 & 0.226 & 0.371\\
 M1.2      & 0.193 & 0.311 & 0.199 & 0.219 & 0.220 & 0.358\\
 \bottomrule
\end{tabular}}
\caption{\label{tab:auto-content}
Results for multilingual formality transfer on content preservation using one reference.
}
\end{table}

\paragraph{Direction Analysis}

For English,~\citet{rao-tetreault-2018} find that the I$\rightarrow$F direction is quite different from the opposite one since there are far more ways to express informality. 
As our work is the first attempt at the F$\rightarrow$I direction in a multilingual setting, 
we run some additional analysis 
using two test sets for each direction: (a) the original
test set; (b) the test set of the
opposite direction, swapping sources and references. 
We fine-tune BART~\citep{lewis-etal-2020-bart} and mBART-50~\citep{tang2020multilingual} with English parallel data (GYAFC) and evaluate them on (a) and (b).
Figure~\ref{fig:en-content} shows the results of content preservation. For INPUT (source copy), BLEU scores are almost the same swapping sources and references but COMET ones are not, probably due to COMET being trained to prefer a formal/better ``generated sentence''; 
compared to INPUT, the performance gain of BART and mBART in I$\rightarrow$F is larger than the opposite direction on both  metrics. Results are similar for other languages (Table~\ref{tab:auto-content}). We pick M1.1 and M1.2 from Table~\ref{tab:auto-results} since they are both fine-tuned using parallel data in the target language. BLEU scores of F$\rightarrow$I are always lower than the opposite; the COMET score of INPUT in F$\rightarrow$I is higher than I$\rightarrow$F, but scores of both systems for F$\rightarrow$I drop after transforming the source sentence into the target style. All these observations suggest that there is more variation in informal texts for the languages we consider, and the F$\rightarrow$I direction is harder.

\section{Conclusions}
Fine-tuning a pre-trained multilingual model with machine translated training data 
yields state-of-the-art results for transferring informal to formal text.
The results for the formal-to-informal direction are considerably worse---the task is more difficult, and the quality of translated informal text is lower.
We have also proposed two adaptation training strategies that can be applied in a cross-lingual transfer strategy 
. These strategies target language and task adaptation, and can be combined to adapt mBART for multilingual formality transfer. The adaptation strategies with auxiliary parallel data from a different language are effective, yielding competitive results and outperforming more classic IBT-based approaches without task-specific parallel data. 
Lastly, we have shown that formal-to-informal transformation is harder than the opposite direction.

\section*{Acknowledgments}

This work was partly funded by the China Scholarship Council (CSC). The anonymous reviewers of ACL Rolling Review provided us with useful comments which contributed to improving this paper and its presentation, so we're grateful to them. We would also like to thank the Center for Information Technology of the University of Groningen for their support and for providing access to the Peregrine high performance computing cluster.

\section*{Ethics Statement}

All work that automatically generates and/or alters natural text could unfortunately be used maliciously. While we cannot fully prevent such uses once our models are made public, we do hope that writing about risks explicitly and also raising awareness of this possibility in the general public are ways to contain the effects of potential harmful uses. We are open to any discussion and suggestions to minimise such risks.

\bibliography{anthology,custom}
\bibliographystyle{acl_natbib}

\clearpage
\appendix
\onecolumn
\section{\Large Appendices: \\~ \\ }
\label{sec:appendix}

\setcounter{table}{0}
\renewcommand{\thetable}{A.\arabic{table}}
\setcounter{figure}{0}
\renewcommand{\thefigure}{A.\arabic{figure}}

\bigskip

This appendices include: (i) Results for BART and mBART on English data (\ref{app:bart-mbart}); (ii) Results for style classifiers/regressor (\ref{app:class-regr}); (iii) Detailed results for multilingual formality transfer (\ref{app:all-results})\vspace*{0.5cm}. 

\subsection{Results for BART and mBART on English data}
\label{app:bart-mbart}

We fine-tune BART~\citep{lewis-etal-2020-bart} and mBART-50~\citep{tang2020multilingual} with English parallel data specifically developed for formality transfer in English (GYAFC). The performance of  BART and English data can be seen as a sort of upperbound, as these are best conditions (monolingual model, and gold parallel data). The drop we see using mBART is rather small, suggesting mBART is a viable option. We also see that formal to informal is much harder than viceversa, probably due to high variability in informal formulations~\citep{rao-tetreault-2018}. Table~\ref{tab:results-bart-mbart} shows the results for both models.

\begin{table}[ht]
\centering
\small
\begin{tabular}{lcccccc}
\toprule
\sc Model & \sc Direction & COMET & BLEU & REG. & ACC & HM\\
\hline
\multirow{2}{*}{BART} & Informal$\rightarrow$Formal & 0.544 & 0.795 & -0.527 & 0.928 & 0.856\\
                      & Formal$\rightarrow$Informal & 0.170 & 0.436 & -1.143 & 0.683 & 0.532\\
\hline
\multirow{2}{*}{mBART}& Informal$\rightarrow$Formal & 0.512 & 0.779 & -0.531 & 0.916 & 0.842\\
                      & Formal$\rightarrow$Informal & 0.151 & 0.422 & -1.031 & 0.591 & 0.492\\
\bottomrule
\end{tabular}
\caption{\label{tab:results-bart-mbart}
Results of BART and mBART on English data. Note that REG. indicates the score of the style regressor (the higher is better in Informal$\rightarrow$Formal, lower is better in Formal$\rightarrow$Informal).
}
\end{table}

\subsection{Results for style classifiers/regressor}
\label{app:class-regr}
We compare four different style classifiers and one regressor: (i) TextCNN~\citep{kim-2014-convolutional} trained with pseudo-parallel data in the target language; (ii) mBERT~\citep{devlin-etal-2019-bert} fine-tuned with pseudo-parallel data, English data, or a combination of all data; and (iii) a XLM-R~\citep{conneau-etal-2020-unsupervised} based style regressor from~\citet{briakou-etal-2021-evaluating}, which is trained with formality rating data in English.

\begin{table*}[ht]
\centering
\resizebox{\linewidth}{!}{%
\begin{tabular}{lccccccccccccc}
\toprule[1pt]
 \multirow{2}{*}{\sc Model} & \multirow{2}{*}{\sc Training Data} & \multicolumn{4}{c}{\sc Italian} & \multicolumn{4}{c}{\sc French} & \multicolumn{4}{c}{\sc Portuguese}\\ 
 \cline{3-14}
 & & ACC & Precision & Recall & F1 & ACC & Precision & Recall & F1 & ACC & Precision & Recall & F1 \\
\toprule[0.6pt]
 TextCNN & Pseudo data & 0.865 & 0.885 & 0.839 & 0.861 & 0.838 & 0.876 & 0.787 & 0.829 & 0.799 & 0.793 & 0.809 & 0.801\\
 mBERT & Pseudo data & \textbf{0.898} & 0.905 & 0.890 & \textbf{0.897} & 0.879 & \textbf{0.918} & 0.831 & 0.872 & \textbf{0.851} & 0.806 & 0.924 & \textbf{0.861}\\
 mBERT & English data & 0.889 & 0.856 & \textbf{0.934} & 0.893 & \textbf{0.896} & 0.856 & \textbf{0.951} & \textbf{0.901} & 0.839 & 0.771 & \textbf{0.964} & 0.857\\
 mBERT & All data & 0.891 & \textbf{0.906} & 0.872 & 0.888 & 0.882 & 0.911 & 0.846 & 0.877 & \textbf{0.851} & \textbf{0.815} & 0.909 & 0.859\\
 \hline
 XLM-R & Formality ratings & \multicolumn{2}{c}{Informal: -1.672} & \multicolumn{2}{c}{Formal: 0.108} & \multicolumn{2}{c}{Informal: -1.701} & \multicolumn{2}{c}{Formal: 0.050} & \multicolumn{2}{c}{Informal: -1.438} & \multicolumn{2}{c}{Formal: 0.065} \\
\bottomrule[1pt]
\end{tabular}}
\caption{\label{tab:style-results}
Results for style classifiers/regressor on test set. The data used for evaluation are 1000 sentences from the test set and the corresponding 1000 human references. For informal sentences, the smaller the XLM-R score is better, higher is better for formal sentences.
}
\end{table*}

\clearpage
\subsection{Detailed results for multilingual formality transfer}
\label{app:all-results}
\begin{table*}[ht]
\centering
\resizebox{\linewidth}{!}{%
\begin{tabular}{c|l|ccccc|ccccc|ccccc}
\toprule[1.5pt]
\makecell[c]{\multirow{2}{*}{\sc Data}} & \makecell[c]{\multirow{2}{*}{\sc Model}} & \multicolumn{5}{c|}{\textsc{Italian}} & \multicolumn{5}{c|}{\textsc{French}} & \multicolumn{5}{c}{\textsc{Portuguese}}\\
  \cline{3-17}
  & & COMET & BLEU & REG. & ACC & HM & COMET & BLEU & REG. & ACC & HM & COMET & BLEU & REG. & ACC & HM \\
  \hline
  \rowcolor{LightGray}
 \multicolumn{17}{c}{\sc Transfer Direction: Informal$\rightarrow$Formal}\\

 \hline
  \multirow{6}{*}{D1} & Translate Train Tag~\citep{briakou-etal-2021-ola}   & -0.059 & 0.426 & -0.705 & 0.735 & 0.539 & -0.164 & 0.451 & -0.586 & 0.696 & 0.547 & 0.194 & 0.524 & -0.636 & 0.755 & 0.619\\
  & + Back-Tranlated Data~\citep{briakou-etal-2021-ola} & 0.026 & 0.430 & -0.933 & 0.556 & 0.485 & 0.004 & 0.491 & -0.898 & 0.485 & 0.488 & 0.301 & 0.546 & -0.875 & 0.627 & 0.584\\
  & Multi-Task Tag-Style~\citep{briakou-etal-2021-ola}  & -0.021 & 0.426 & -0.698 & 0.727 & 0.537 & -0.062 & 0.480 & -0.501 & 0.742 & 0.583 & 0.266 & 0.550 & -0.578 & 0.782 & 0.645\\
  & M1.1: pseudo-parallel data                        & 0.143 & 0.459 & -0.426 & \underline{\textbf{0.856}} & \underline{\textbf{0.598}} & 0.124 & \textbf{0.530} & -0.305 & 0.829 & 0.647 & 0.297 & 0.524 & \underline{\textbf{-0.334}} & \textbf{0.852} & 0.649\\
  & M1.2: M1.1 + EN parallel data                     & \underline{\textbf{0.147}} & \textbf{0.461} & -0.442 & 0.841 & 0.596 & \underline{\textbf{0.130}} & 0.525 & -0.275 & \underline{\textbf{0.863}} & \underline{\textbf{0.653}} & \underline{\textbf{0.331}} & \underline{\textbf{0.553}}  & -0.395 & 0.809 & \underline{\textbf{0.657}}\\
  & M1.3: all data (one model)                        & 0.137 & \textbf{0.461} & \underline{\textbf{-0.409}} & 0.850 & \underline{\textbf{0.598}} & 0.127 & 0.515 & \underline{\textbf{-0.267}} & 0.851 & 0.642 & 0.309 & 0.537 & -0.367 & 0.803 & 0.644\\

  \hline
  \multirow{7}{*}{D2} & DLSM~\citep{briakou-etal-2021-ola}                  & -1.332 & 0.124 & -2.141 & 0.223 & 0.159 & -1.267 & 0.180 & -2.021 & 0.152 & 0.165 & -1.131 & 0.185 & -2.078 & 0.191 & 0.188\\
  & M2.1: IBT training                    & 0.057 & 0.420 & -1.351 & 0.240 & 0.305 & -0.019 & 0.465 & -1.303 & 0.219 & 0.298 & 0.233 & 0.487 & -1.074 & 0.411 & 0.446\\
  & M2.2: M2.1 + EN data                  & 0.105 & 0.460 & -0.867 & 0.510 & 0.484 & 0.036 & 0.500 & -0.814 & 0.487 & 0.492 & 0.236 & 0.491 & -1.040 & 0.428 & 0.457\\
  & M2.3: ADAPT + EN  cross-attn   & \textbf{0.139} & 0.467 & -0.684 & 0.637 & 0.539 & 0.066 & 0.516 & -0.613 & 0.627 & 0.566 & 0.288 & 0.499 & -1.143 & 0.365 & 0.422\\
  & M2.4: ADAPT + EN data       & 0.131   & \textbf{0.476} & \textbf{-0.537} & \textbf{0.731} & \textbf{0.577} & \textbf{0.074} & \textbf{0.519} & \textbf{-0.572} & \textbf{0.702} & \textbf{0.597} & \textbf{0.291} & \textbf{0.526} & \textbf{-0.922} & \textbf{0.509} & \textbf{0.517}\\

  \hline
  \multirow{3}{*}{D3} & M3.1: EN data                & \textbf{0.134} & \underline{\textbf{0.485}} & -0.590 & 0.670 & \textbf{0.563} & \textbf{0.102} & \underline{\textbf{0.553}} & -0.591 & 0.727 & 0.628 & -1.673 & 0.039 & \textbf{-0.550} & \underline{\textbf{0.890}} & 0.074\\
  & M3.2: ADAPT + EN cross-attn  & 0.130 & 0.480 & -0.588 & 0.672 & 0.560 & 0.091 & 0.545 & \textbf{-0.446} & \textbf{0.749} & \textbf{0.631} & \textbf{0.302} & \textbf{0.547} & -0.859 & 0.559 & \textbf{0.553}\\
  & M3.3: ADAPT + EN data       & -0.107 & 0.423 & \textbf{-0.579} & \textbf{0.735} & 0.537 & 0.101 & 0.547 & -0.488 & 0.722 & 0.622 & -0.260 & 0.423 & -1.112 & 0.508 & 0.462\\


  \hline
  \multirow{4}{*}{D4} & Round-trip MT~\citep{briakou-etal-2021-ola} & -0.053 & 0.346 & \textbf{-1.026} & \textbf{0.354} & \textbf{0.350} & -0.065 & 0.416 & \textbf{-0.748} & \textbf{0.406} & \textbf{0.411} & 0.213 & 0.430 & \textbf{-0.661} & \textbf{0.601} & \textbf{0.501}\\
  & Rule-based~\citep{briakou-etal-2021-ola} & \textbf{0.071} & \textbf{0.438} & -1.167 & 0.268 & 0.333 & \textbf{-0.013} & \textbf{0.472} & -1.236 & 0.208 & 0.289 & \textbf{0.291} & \textbf{0.535} & -1.081 & 0.448 & 0.488\\
  & M4.1: original mBART & -0.067 & 0.380 & -1.672 & 0.103 & 0.162 & -0.106 & 0.425 & -1.709 & 0.080 & 0.135 & -1.444 & 0.128 & -1.870 & 0.200 & 0.156\\
  & M4.3: ADAPT (generic data) & 0.033 & 0.401 & -1.675 & 0.092 & 0.150 & -0.033 & 0.444 & -1.700 & 0.075 & 0.128 & 0.230 & 0.463 & -1.438 & 0.223 & 0.301\\
  \hline
  \rowcolor{LightGray}
  \multicolumn{17}{c}{\sc Transfer Direction: Formal$\rightarrow$Informal}\\ 

 \hline
  \multirow{3}{*}{D1} &  M1.1: pseudo-parallel data            & \textbf{0.298} & 0.177 & -0.225 & 0.311 & 0.226 & \textbf{0.239} & \textbf{0.195} & -0.188 & 0.377 & 0.257 & 0.388 & 0.225 & -0.273 & 0.306 & \textbf{0.259}\\
  & M1.2: M1.1 + EN parallel data         & 0.278 & \textbf{0.178} & -0.228 & 0.315 & 0.227 & 0.215 & 0.194 & \textbf{-0.304} & \textbf{0.458} & \textbf{0.273} & 0.373 & 0.219 & \textbf{-0.282} & \textbf{0.313} & 0.258\\
  & M1.3: all data (one model)            & 0.283 & 0.175 & \textbf{-0.287} & \textbf{0.368} & \textbf{0.237} & 0.207 & 0.191 & -0.301 & 0.439 & 0.266 & \textbf{0.407} & \textbf{0.229} & -0.241 & 0.292 & 0.257\\
  
  \hline
  \multirow{4}{*}{D2} &  M2.1: IBT training& 0.335 & 0.166 & -0.082 & 0.338 & 0.223 & 0.272 & 0.195 & 0.037 & 0.194 & 0.194 & 0.467 & \underline{\textbf{0.237}} & 0.042 & 0.084 & 0.124\\
  & M2.2: M2.1 + EN data                  & \underline{\textbf{0.337}} & 0.168 & -0.174 & 0.420 & 0.240 & \underline{\textbf{0.274}} & 0.196 & -0.016 & 0.235 & 0.214 & \textbf{0.471} & \underline{\textbf{0.237}} & 0.045 & 0.083 & 0.123 \\
  & M2.3: ADAPT + EN cross-attn  & 0.176 & 0.175 & \textbf{-0.631} & 0.672 & 0.278 & 0.226 & \textbf{0.212} & \textbf{-0.464} & \textbf{0.627} & \textbf{0.317} & 0.441 & \underline{\textbf{0.237}} & -0.343 & 0.471 & \underline{\textbf{0.315}}\\
  & M2.4: ADAPT + EN data       & 0.279 & \textbf{0.180} & -0.582 & \textbf{0.719} & \textbf{0.288} & 0.232 & 0.209 & -0.444 & 0.567 & 0.305 & -0.022 & 0.169 & \textbf{-0.520} & \textbf{0.534} & 0.257\\
  \hline
  \multirow{3}{*}{D3} & M3.1: EN data                & 0.289 & \underline{\textbf{0.186}} & -0.646 & \underline{\textbf{0.767}} & \underline{\textbf{0.299}} & \textbf{0.244} & \underline{\textbf{0.216}} & -0.566 & \underline{\textbf{0.692}} & \underline{\textbf{0.329}} & -1.695 & 0.020 & \textbf{-1.225} & 0.403 & 0.038\\
  & M3.2: ADAPT + EN cross-attn  & \textbf{0.300} & 0.179 & -0.285 & 0.421 & 0.251 & 0.221 & 0.209 & \underline{\textbf{-0.594}} & 0.685 & 0.320 & \textbf{0.367} & 0.175 & -0.449 & \textbf{0.560} & 0.267\\
  & M3.3: ADAPT + EN data       & 0.100 & 0.169 & \underline{\textbf{-0.744}} & 0.733 & 0.275 & 0.220 & 0.205 & -0.447 & 0.584 & 0.303 & 0.130 & \textbf{0.189} & -0.586 & 0.505 & \textbf{0.275}\\
  

  \hline
  \multirow{2}{*}{D4} & M4.1: original mBART & 0.260 & 0.160 & \textbf{0.076} & \textbf{0.146} & \textbf{0.153} & 0.204 & 0.189 & \textbf{0.031} & \textbf{0.189} & \textbf{0.189} & -1.363 & 0.080 & \underline{\textbf{-1.406}} & \underline{\textbf{0.657}} & \textbf{0.143}\\
  & M4.2: ADAPT (generic data) & \textbf{0.317} & \textbf{0.164} & 0.084 & 0.130 & 0.145 & \textbf{0.268} & \textbf{0.194} & 0.052 & 0.170 & 0.181 & \underline{\textbf{0.475}} & \underline{\textbf{0.237}} & 0.047 & 0.082 & 0.122\\
 \bottomrule[1.5pt]
\end{tabular}}
\caption{\label{tab:all-results}
Results for multilingual formality transfer. Notes: (i) REG. indicates the score of the style regressor (the higher is better in I$\rightarrow$F, lower is better in F$\rightarrow$I); (ii) for F$\rightarrow$I there are four different source sentences and a human reference only, so for each instance scores are averaged; (iii) bold numbers denote best systems for each block, and underlined indicate the best score for each transfer direction.
}
\end{table*}

\end{document}